\documentclass{article}
\usepackage{ijcai18}

\usepackage{times}
\usepackage{xcolor}
\usepackage{soul}
\usepackage[utf8]{inputenc}
\usepackage[small]{caption}

\usepackage{url}
\usepackage{graphicx}
\usepackage{booktabs}
\usepackage{amsfonts}
\usepackage{nicefrac}
\usepackage{microtype}
\usepackage{amsmath}
\usepackage{algorithm}
\usepackage{algpseudocode}
\usepackage{tikz}
\usepackage{xspace}
\usepackage{subcaption}
\usepackage{pgfplots}
\usepackage{arydshln}
\usepackage{multirow}
\usepackage{placeins}

\newcommand{\citet}[1]
{\citeauthor{#1}~\shortcite{#1}}

\algtext*{EndWhile}% Remove "end while" text
\algtext*{EndIf}% Remove "end if" text
\algtext*{EndFor}% Remove "end if" text

\usetikzlibrary{trees}
\usetikzlibrary{arrows,shapes,shapes.geometric,backgrounds}
\usetikzlibrary{calc}
\usetikzlibrary{decorations.markings}
\usetikzlibrary{fit}

\tikzstyle{bnarrow}=[
    decoration={markings,mark=at position 1 with {\arrow[scale=1.5]{>}}},
    postaction={decorate},
    shorten >=0.7pt,
    >=latex,
    line width=0.3
]
\tikzstyle{bayesnet}=[
  >=latex, thick, auto
]
\tikzstyle{bnnode}=[
  draw,ellipse,minimum size=7mm,inner sep=1pt,font=\small
]
\tikzstyle{circnode}=[
  draw,circle,minimum size=7mm,inner sep=1pt,font=\small
]
\tikzstyle{cpt}=[
  font=\footnotesize
]

\makeatletter

\let\c@table\c@figure
\makeatother

\newcommand\given[1][]{\:#1\vert\:}

\DeclareMathOperator{\SDP}{SDP}
\DeclareMathOperator{\ECA}{ECA}
\DeclareMathOperator{\MPA}{MPA}

\DeclareMathOperator{\MAA}{MAA}

\DeclareMathOperator*{\argmax}{arg\,max}

\newcommand{\rvars}[1]{\ensuremath{\mathbf{#1}}\xspace}
\newcommand{\Xs}{\rvars{X}}
\newcommand{\Ys}{\rvars{Y}}
\newcommand{\Zs}{\rvars{Z}}
\newcommand{\Es}{\rvars{E}}
\newcommand{\Fs}{\rvars{F}}
\newcommand{\Gs}{\rvars{G}}

\newcommand{\Rs}{\rvars{R}}
\newcommand{\Qs}{\rvars{Q}}
\newcommand{\Is}{\rvars{I}}

\newcommand{\jstate}[1]{\ensuremath{\mathbf{#1}}\xspace}
\newcommand{\xs}{\jstate{x}}
\newcommand{\ys}{\jstate{y}}
\newcommand{\zs}{\jstate{z}}
\newcommand{\es}{\jstate{e}}
\newcommand{\fs}{\jstate{f}}
\newcommand{\gs}{\jstate{g}}

\newcommand{\rs}{\jstate{r}}

\newtheorem{prop}{Proposition}
\newtheorem{defn}{Definition}
\newtheorem{cor}{Corollary}
\newtheorem{claim}{Claim}

\makeatletter
\DeclareRobustCommand{\qed}{%
  \ifmmode % if math mode, assume display: omit penalty etc.
  \else \leavevmode\unskip\penalty9999 \hbox{}\nobreak\hfill
  \fi
  \quad\hbox{\qedsymbol}}
\newcommand{\openbox}{\leavevmode
  \hbox to.77778em{%
  \hfil\vrule
  \vbox to.675em{\hrule width.6em\vfil\hrule}%
  \vrule\hfil}}
\newcommand{\qedsymbol}{\openbox}
\newenvironment{proof}[1][\proofname]{\par
  \normalfont
  \topsep6\p@\@plus6\p@ \trivlist
  \item[\hskip\labelsep\itshape
    #1.]\ignorespaces
}{%
  \qed\endtrivlist
}
\newcommand{\proofname}{Proof}
\makeatother

\newcommand{\eat}[1]{}

\newcommand{\yj}[1]{\textcolor{cyan}{\textbf{[YooJung: #1]}}}

\title{On Robust Trimming of Bayesian Network Classifiers}
\author{YooJung Choi \and Guy Van den Broeck\\ 
Computer Science Department\\
University of California, Los Angeles  \\
\{yjchoi, guyvdb\}@cs.ucla.edu}

\begin{document}
\maketitle

\begin{abstract}
This paper considers the problem of removing costly features from a Bayesian network classifier. We want the classifier to be robust to these changes, and maintain its classification behavior. To this end, we propose a closeness metric between Bayesian classifiers, called the {\em expected classification agreement (ECA).}
Our corresponding trimming algorithm finds an optimal subset of features and a new classification threshold that maximize the expected agreement, subject to a budgetary constraint. It utilizes new theoretical insights to perform branch-and-bound search in the space of feature sets, while computing bounds on the ECA.
Our experiments investigate both the runtime cost of trimming and its effect on the robustness and accuracy of the final~classifier.
\end{abstract}

\section{Introduction} \label{s:intro}

% SETTING
Bayesian classification plays a prominent role throughout machine learning~\cite{wu2008top,laidlaw1998partial,metsis2006spam}.
In this setting, one has a model that specifies a probability distribution $\Pr$ over a set of variables, including class variable $C$ and attributes or features $\Fs = \{F_1, \ldots, F_n\}$.
Given a particular instance, described as an assignment to features $\fs=\{f_1, \ldots, f_n\}$, this model is used to compute the posterior probability $\Pr(C | f_1, \ldots, f_n)$ which is then compared against a threshold $T$ to classify the~instance.

% WHY FEATURE SELECTION?
In practice, observing features often has a cost, and one typically needs to keep it within a given budget.
For example, features in a medical diagnosis may be invasive, time-consuming, or expensive medical tests~\cite{kononenko2001machine}.
Similar issues arise in active sensing~\cite{Gao11}, adaptive testing~\cite{millan2002,munie2008}, and robotics~\cite{kollar2008efficient}.
This problem has been studied from different angles, often under the umbrella of {\em feature selection}.
For example, one may select features at learning time based on their relevance, redundancy, or classification accuracy~\cite{kira1992feature,yu2004efficient}. 
Alternatively, features may be selected at prediction time based on their expected misclassification cost or information gain~\cite{bilgic11,krause2009,Zhang2010}. Such probabilistic objectives are computed on the distribution of the Bayesian classifier.

% WHY ROBUST FEATURE SELECTION?
This paper approaches the problem from a different perspective, which we call classifier \emph{trimming}.
In addition to selecting features that fit the budget, trimming adjusts the threshold $T$ to induce a new classifier.
Moreover, instead of simply optimizing the predictive accuracy, we want trimming to be robust. That is, we want to preserve the original classifier's \emph{general behavior}.
Our motivation is two-fold. 
First, Bayesian classifiers often incorporate significant expert knowledge in the form of priors, structural assumptions, and choice of distribution class~\cite{lucas2001expert}. This is particularly true in medical applications where data is scarce~\cite{bellazzi2008predictive}.
Second, two classifiers with the same predictive quality can exhibit vastly different behavior and failure modes. For example, \citeauthor{BCWS16}~\shortcite{BCWS16} describe two classifiers with a similar accuracy, but markedly different amounts of gender bias in their predictions.
In either scenario, it is essential to retain the desired behavior of the original classifier during trimming.

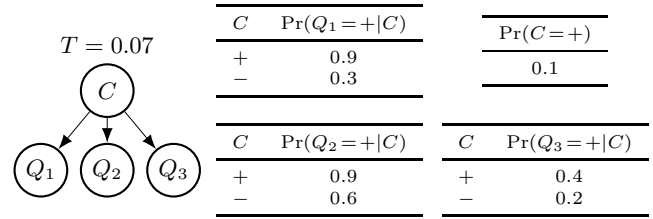
\begin{figure}[tb]
    \centering
    % \scalebox{0.9}{
      \begin{tikzpicture}[bayesnet]
      
  \def\lone{-55bp}
  \def\ltwo{-85bp}

  \node [label={above:\small$T=0.07$}] (C) at (-80bp,\lone) [bnnode] {$C$};
  \node (Q1) at (-105bp,\ltwo) [bnnode] {$Q_1$};
  \node (Q2) at (-80bp,\ltwo) [bnnode] {$Q_2$};
  \node (Q3) at (-55bp,\ltwo) [bnnode] {$Q_3$};
  
  \begin{scope}[on background layer]
    \draw [bnarrow] (C) -- (Q1);
    \draw [bnarrow] (C) -- (Q2);
    \draw [bnarrow] (C) -- (Q3);
  \end{scope}
  
  \node at (85bp,-40bp) [cpt] {
    \scriptsize
    \begin{tabular}{c}
      \toprule
      $\Pr(C\!=\!+)$\\\midrule
      $0.1$\\
      \bottomrule
    \end{tabular}
  };
  
  \node at (0bp,-40bp) [cpt] {
    \scriptsize
    \begin{tabular}{lc}
      \toprule
      $C$ & $\Pr(Q_1\!=\!+|C)$\\\midrule
      $+$ & $0.9$\\
      $-$ & $0.3$\\
      \bottomrule
    \end{tabular}
  };
  
  \node at (0bp,-85bp) [cpt] {
    \scriptsize
    \begin{tabular}{lc}
      \toprule
      $C$ & $\Pr(Q_2\!=\!+|C)$\\\midrule
      $+$ & $0.9$\\
      $-$ & $0.6$\\
      \bottomrule
    \end{tabular}
  };
  
  \node at (85bp,-85bp) [cpt] {
    \scriptsize
    \begin{tabular}{lc}
      \toprule
      $C$ & $\Pr(Q_3\!=\!+|C)$\\\midrule
      $+$ & $0.4$\\
      $-$ & $0.2$\\
      \bottomrule
    \end{tabular}
  };
    \end{tikzpicture}
    % }
\caption{Naive Bayes classifier for a quiz scenario where answers on $\Qs=\{Q_1,Q_2,Q_3\}$ (features) depend on knowledge $C$ (class)}\label{fig:nb}
\end{figure}

Figure~\ref{fig:nb} depicts a classifier utilizing three features $\Qs=\{Q_1,Q_2,Q_3\}$ with a threshold of $T=0.07$.
Consider two possible trimmings of this classifier: one obtained by removing $Q_2$ and adjusting the threshold to $0.10$, the other with $Q_1$ removed and the threshold changed to $0.30$.
The trimmed classifiers are clearly less expensive than the original one, but how do we quantitatively compare and choose between these trimmings?
To answer this question, we introduce the notion of {\em expected classification agreement (ECA)}. It is an expectation of the two classifiers agreeing on instances, measuring how much behavior from the original classifier is preserved.

Probabilistic graphical models, such as Bayesian networks, are often used to represent the  Bayesian classifier's distribution. 
We propose an algorithm to find the best trimming of a Bayesian network classifier subject to a budgetary constraint. The algorithm selects features and chooses a new classification threshold in order to maximize the ECA.
We also propose a specialized algorithm for the case of naive Bayes classifiers~\cite{friedman1997bayesian,cheng1999comparing} that exploits the naive Bayes independence assumptions for more efficient trimming.
These novel trimming algorithms are based on the following progression of ideas. 
First, we show how an existing compilation algorithm to compute {\em expected same-decision probability (E-SDP)} can be modified to compute the ECA between a classifier and its trimming~\cite{ChoiIJCAI17}. This objective was previously used for feature selection where the classification threshold remains fixed~\cite{chen15}.
Second, we propose an upper bound on the ECA that can be computed more efficiently, enabled by our formulation that adjusts the threshold.
Lastly, we use this upper bound to effectively trim classifiers with branch-and-bound search.

Finally, with evaluation on real-world data, we show that our approach finds robust trimmings and demonstrate the relationship between robustness and accuracy.
We also illustrate the importance of optimizing the threshold for both classification similarity and efficiency of search.
Moreover, we show that our trimming approach consistently returns a classifier that is significantly more similar to the original classifier than selecting features based on information gain.

% This paper is structured as follows. First, we introduce the problem more formally and define the new notion of expected classification agreement.
% Next, we describe our approach to searching for optimal trimmings of Bayesian network classifiers. We define the upper bound we use in the search algorithm, and introduce its desirable theoretical properties.
% We then show a computational connection between the ECA and the (expected) same-decision probability, and present algorithms that exploit this connection to compute the ECA and its upper bound.
% We follow with experimental results and finally close with concluding remarks.
% \guy{This last paragraph can be removed for a short 6-page paper.}

\section{Expected Classification Agreement}\label{sec:eca}

We use the standard notation where variables are denoted by upper case letters ($X$) and their instantiations by lower case letters ($x$). Sets of variables are denoted in bold upper case ($\Xs$) and their joint instantiations in bold lower case ($\xs$). Concatenations of sets ($\Xs\Ys$) represent their union.

A binary Bayesian classifier is a tuple $\alpha = (C,\Fs,T)$, where \(C\)  is a binary class variable, $\Fs$ are (possibly multi-valued) features, and  $T$ is a threshold.
On a joint probability distribution $\Pr(.)$ over variables $C$ and $\Fs$, the classification function is
\begin{align*}
    C_T(\fs) =
    \begin{cases}
        c, & \text{if} \Pr(c \given \fs) \geq T \\
        \overline{c}, & \text{otherwise.}
    \end{cases}
\end{align*}
% That is, an instance is classified into a particular class $c$ iff $\Pr(c | f_1, \ldots, f_n) \geq T$.
For example, with a threshold of $0.5$, an instance will be classified into the more probable class after observing its features.

Next, we motivate and define our proposed closeness measure between classifiers, quantifying their expected agreement.

\subsection{Example and Motivation}

Consider again the scenario shown in Figure~\ref{fig:nb}, where an instructor uses a quiz to test students' knowledge.
The quiz contains three independent questions: $Q_1$ is strongly indicative of being knowledgeable,
$Q_2$ is an easy question, and
$Q_3$ is a hard question (only 40\% of the knowledgeable students answer it correctly).
The subject of this quiz is quite difficult, and only 10\% of the students are expected to master it, as reflected by the prior on class variable \(C\).
Hence, the instructor sets a lenient threshold of $T = 0.07$ to avoid failing students who may have grasped the subject.

According to this classifier, a student will pass the quiz precisely when their answer matches one of the following three (out of eight)
outcomes: $\{Q_1\!=\!+, Q_2\!=\!+, Q_3\!=\!+\}$, $\{Q_1\!=\!+, Q_2\!=\!+, Q_3\!=\!-\}$,  and $\{Q_1\!=\!+, Q_2\!=\!-, Q_3\!=\!+\}$.
Moreover, the probability of seeing one of these outcomes is $32\%$: the fraction of students that are expected to pass the quiz.
Suppose now that we drop questions \(Q_1\) and \(Q_2\), relying solely on question $Q_3$ to evaluate students (using the same threshold).
Since \(\Pr(C\!=\!+ \given Q_3)\) is always greater than \(T = 0.07\), all students will pass the quiz, completely ignoring the test results.
Alternatively, we can make more intuitive use of the test question and pass only the students who answered $Q_3$ correctly. This is equivalent to comparing $\Pr(C\!=\!+ \given Q_3)$ against a new threshold of $T = 0.15$.
Using this new threshold, we will now obtain the same student assessment on five test outcomes,\footnote{The two classifiers will disagree on $\{Q_1\!=\!+, Q_2\!=\!+, Q_3\!=\!-\}$, $\{Q_1\!=\!-, Q_2\!=\!+, Q_3\!=\!+\}$ and $\{Q_1\!=\!-, Q_2\!=\!-, Q_3\!=\!+\}$.}
whose probabilities add up to $75\%$. This is the expected classification agreement (ECA). In particular, we say that the two classifiers
$\alpha = (C,\{Q_1,Q_2,Q_3\},0.07)$ and $\beta = (C,\{Q_3\},0.15)$ have an ECA of~$75\%$.

\subsection{Formalization}

We now formalize the notion of ECA and classifier trimming.

\begin{defn}
Let $\alpha = (C,\Fs,T)$ be a Bayesian classifier using distribution \(\Pr(.)\). The classifier $\beta = (C,\Fs^\prime,T^\prime)$ is a \emph{trimming} of \(\alpha\) if it uses
the same class variable $C$ and distribution \(\Pr(.)\) as \(\alpha\), and a subset of its features (i.e., \(\Fs^\prime \subset \Fs\)).
\end{defn}

\begin{defn}
Let $\alpha = (C,\Fs,T)$ be a Bayesian classifier and let $\beta = (C,\Fs^\prime,T^\prime)$ be one of its trimmings. The \emph{expected classification agreement (ECA)} between these classifiers is:
\begin{align*}
  \ECA(\alpha,\beta)
  = \sum_{\fs} \left[C_T(\fs) = C_{T^\prime}(\fs^\prime)\right] \cdot \Pr(\fs).
\end{align*}
Here, $\fs^\prime$ is the subset of instantiation $\fs$ pertaining to variables in $\Fs^\prime$, and \([.]\) is an indicator function (evaluates to \(1\) when its argument is true and to \(0\) otherwise).
\end{defn}
Section~\ref{s:intro} asks to compare trimmings of classifier $\alpha$ in Figure~\ref{fig:nb}. The first trimming has \(\ECA(\alpha,(C,\{Q_1,Q_3\},0.10)) = 91\%\) while the second has \(\ECA(\alpha,(C,\{Q_2,Q_3\},0.30)) = 68\%\).

We are now ready to define the {\em classifier trimming problem} more formally.
The input to this problem is a binary Bayesian classifier $\alpha = (C,\Fs,T)$, a positive cost for each feature in $\Fs$, and a budget $B$.
The output is a subset of features $\Fs^{\star} \subseteq \Fs$ whose sum of costs is at most $B$ and a threshold $T^\star$, leading to a trimmed classifier $\beta^\star = (C,\Fs^{\star},T^{\star})$ that maximizes the ECA with $\alpha$:
\begin{equation*}
  \beta^\star = \argmax_\beta \ECA(\alpha,\beta).
\end{equation*}
In other words, we wish to find a solution to the following optimization problem:
\begin{align}
  \ECA^\star = \max_{\Fs^\prime \subseteq \Fs}\,\max_{T^\prime}\ &\ECA(\alpha, (C,\Fs^\prime,T^\prime))  \nonumber \\
  \text{s.t.} &\sum_{F^\prime\in\Fs^\prime}\text{cost}(F^\prime)\leq B \nonumber
\end{align}

This problem can alternatively be described as feature subset selection using the following criterion.
\begin{defn}
Let $\alpha = (C,\Fs,T)$ be a Bayesian classifier. The \emph{maximum achievable agreement (MAA)} for feature subset $\Fs^\prime \subseteq \Fs$ is defined as:
\begin{align*}
  \MAA_\alpha(\Fs^\prime) = \max_{T^\prime} \ECA(\alpha,(C,\Fs^\prime,T^\prime)).
\end{align*}
\end{defn}
The $\MAA_\alpha(\Fs^\prime)$ corresponds to the maximum ECA that is achievable by a trimmed classifier with features \(\Fs^\prime\). 
Hence, the classifier trimming problem reduces to searching for the subset of features that fits within the budget and maximizes the MAA.
We will drop the subscript $\alpha$ when clear from~context.

%%%%%%%%%%%%%%%%%%%%%%%%%%%%%%%%%%%%%%%%%%%%%%%%%%%%%%%%%%%%%%%%

\section{Searching for an Optimal Trimming}

\begin{figure}[tb]
\centering
\scalebox{0.9}{
\begin{tikzpicture}[level distance=1.2cm,
  level 1/.style={sibling distance=4.0cm},
  level 2/.style={sibling distance=3.0cm},
  level 3/.style={sibling distance=2.0cm},
  tcancel/.append style={draw=#1, cross out, inner sep=1pt}]
  \node [circnode,align=center] {$\{\}$\\$\{\}$}
    child {node [bnnode,align=center] {$\{F_1\}$\\$\{\}$}
        child {
            node (N11) [bnnode,align=center, label={[blue,align=center]below:\small$\MAA=0.9$\\Update $M^\star$}] {$\{F_1,F_2\}$\\$\{\}$}
        }
        child {node (N12)[bnnode,align=center, label={[blue,shift={(0.6,0)}]\small$0.98 > M^\star$}] {$\{F_1\}$\\$\{F_2\}$}
            child {node (N121) [bnnode,align=center, label={[blue,align=center]below:\small$\MAA=0.97$\\Update $M^\star$}] {$\{F_1,F_3\}$\\$\{F_2\}$} }
            child {node (N122) [bnnode,align=center, label={[blue,align=center]below:\small$\MAA=0.9$}] {$\{F_1\}$\\$\{F_2,F_3\}$} } 
        }
    }
    child {
        node [bnnode, align=center, label={[red,shift={(0.5,0)}]\small$0.95 < M^\star$}] {$\{\}$\\$\{F_1\}$}
    }
    ;
\end{tikzpicture}
}
\caption{Branch-and-Bound search tree to select a subset with budget $B=1.5$ among features $\{F_1, F_2, F_3, F_4\}$ with costs $\{0.5, 1.0, 1.0, 2.0\}$ respectively. Every node contains a set of included features and a set of excluded features. \label{fig:bnb}}
\end{figure}
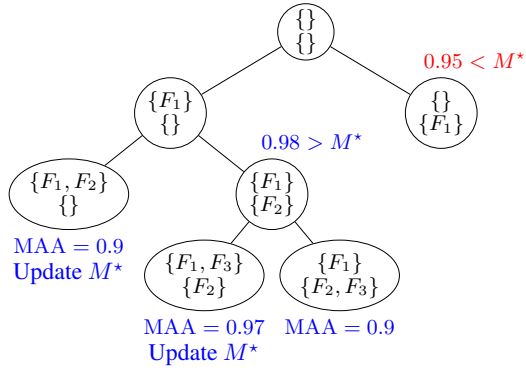

\begin{algorithm}[tb]
\caption{$\textsc{ECA-Trim}(\Is,\Es,b)$}\label{alg:trim-nb}
\textbf{Input:} \\  
  $\alpha$ : Bayesian classifier $(C,\Fs,T)$; \quad $B$ : budget \\
\textbf{Data:} \\
  $\Is \leftarrow \emptyset$, $\Es \leftarrow \emptyset$: set of included/excluded features \\
  $b \leftarrow B$: remaining budget \\
  $\Fs^\star,M^\star,T^\star$ : optimal subset, MAA value, and threshold\\
\textbf{Output:} Optimal trimmed classifier $\beta^\star=(C,\Fs^\star,T^\star)$
\\\hrule
  \begin{algorithmic}[1]
    \If {$b \geq 0$}  \label{line:budget-left}
        \State $(m,T_m) \leftarrow \MAA(\Is)$ \label{line:compute-score}
        \If {$m > M^\star$} $M^\star \leftarrow m$; $\Fs^\star \leftarrow \Is$; $T^\star \leftarrow T_m$ \label{line:update}
        \EndIf
    \EndIf
    \If {$\min_{F\in\Fs\setminus(\Is\cup\Es)}\text{cost}(F) \leq b$}
        \State $m \leftarrow \text{UB}(\Fs\setminus\Es)$
        \If {$m \leq M^\star$} \textbf{return} \label{line:backtrack} \EndIf
        \State $F \leftarrow$ a feature from $\Fs\setminus(\Is\cup\Es)$
        \State $\textsc{ECA-Trim}\big(\Is\cup\{F\},\Es,b-\text{cost}(F)\big)$
        \State $\textsc{ECA-Trim}\big(\Is,\Es\cup\{F\},b\big)$
    \EndIf
  \end{algorithmic}
\end{algorithm}

In this section, we describe our approach to search for an optimal trimming of Bayesian classifiers, or equivalently, selecting a feature subset with optimal MAA.\footnote{Code available at\\ \url{https://github.com/UCLA-StarAI/TrimBN}.} Our approach is based on a branch-and-bound search algorithm similar to~\citet{narendra1977branch} and~\citet{kolesar1967branch}. As shown in Algorithm~\ref{alg:trim-nb}, we run a depth-first search through a binary tree where each node is branched into two nodes: one that includes and one that excludes a feature.
Each node then represents the set of features that are included by the path from the root to that node.
The algorithm computes the MAA at each node if the represented  feature subset fits within the budget, keeping track of the best subset and its MAA at each point in search, as in Lines~\ref{line:budget-left}--\ref{line:update}.
In particular, this means that we compute the MAA even if the subset does not exhaust the budget, because MAA does not necessarily increase as the subset size grows.

The essence of the algorithm is pruning subtrees without affecting the optimality of the solution. That is, the algorithm finds the optimal solution without generating the full binary tree.
Suppose given any node, we know the largest value of MAA that its descendants can achieve (UB).
Then we can safely prune the subtree rooted at that node if the bound does not exceed the current best score. This correspond to backtracking the search, as shown in Line~\ref{line:backtrack} in Algorithm~\ref{alg:trim-nb}.
Formally, let $\Es$ be the set of features that were excluded by the path to a certain node.
Each descendant node will then represent a subset of $\Fs\setminus\Es$.
Hence, if we can compute an upper bound on MAA for all subsets of $\Fs\setminus\Es$, then we can successfully prune intermediate nodes in the search tree.

Figure~\ref{fig:bnb} illustrates an example search execution. The tree is traversed depth-first, from left to right.
If a node represents a feature subset within budget, its MAA is computed, and the score $M^\star$ is updated accordingly.
Otherwise, the upper bound is computed and compared against the current best value.
For example, we backtrack the search after excluding $F_1$, because the upper bound on MAA for subsets of $\{F_2,F_3,F_4\}$ is $0.95$ which is smaller than the current $M^\star=0.97$.
Note that this particular subset has a cost of $4.0$ and does not fit within the budget, but we still compute the bound on it for pruning.
That is, the algorithm computes the upper bound on MAA of the set $\Fs \setminus \Es$ and it subsets, even if their cost may exceed the budget.
We will later show experimentally that the pruning of subtrees makes these extra computations worthwhile. 
\eat{\yj{code: check if a feature is already out of the budget}}

%%%%%%%%%%%%%%%%%%%%%%%%%%%%%%%%%%%%%%%%%%%%%%%%%%%%%%%%%%%%%%%%

\section{Maximum Potential Agreement}

We now introduce an upper bound for the MAA and show how it can be used in the search for an optimal trimming.

\begin{defn}\label{def:mpa}
Consider a Bayesian classifier $\alpha = (C,\Fs,T)$ . Let \(\Fs^{\prime} \subseteq \Fs\) be a subset of its features, and let \(\Rs = \Fs \setminus \Fs^\prime\). The {\em maximum potential agreement} (MPA) is
\begin{align*}
  \MPA_\alpha(\Fs^\prime)
  = \sum_{\fs^\prime} \max_c \left\{\sum_{\rs} \left[C_T(\fs^\prime\rs)=c\right]\cdot\Pr(\fs^\prime\rs) \right\}.
\end{align*}
\end{defn}
Intuitively, the MPA is the expected agreement between a Bayesian classifier \(\alpha\) and a hypothetical classifier \(\gamma\) that classifies an instance $\fs^\prime$ into the class that is more likely after observing the remaining features in $\Rs$.
Note that such classifier $\gamma$ is not a Bayesian classifier as it does not test the posterior $\Pr(c \given \fs^\prime)$ against a threshold.
However, the MPA is still a useful computational tool due to its relationship to the MAA.
\begin{prop}\label{prop:mpa}
The MPA is an upper bound on the MAA: $\MAA_\alpha(\Fs^\prime) \leq \MPA_\alpha(\Fs^\prime)$.
\end{prop}
In addition, the MPA is monotonically increasing, a property that we utilize later in the proposed algorithms.
\begin{prop}\label{prop:mpa-mono}
For any $\Fs_1\!\subseteq\!\Fs_2$, $\MPA_\alpha(\Fs_1) \leq \MPA_\alpha(\Fs_2)$.
\end{prop}
These two propositions together imply that the MPA of $\Fs^\prime$ also upper-bounds the MAA of all subsets of $\Fs^\prime$.
\begin{cor}\label{cor:mpa}
For any $\Fs_1 \subseteq \Fs_2$, $\MAA_\alpha(\Fs_1) \leq \MPA_\alpha(\Fs_2)$.
\end{cor}
Therefore, we can use the MPA as an upper bound on the MAA of a node's descendants in the branch-and-bound search algorithm for optimal trimming.

Lastly, we provide an observation that leads to a computational gain, especially in the case of naive Bayes models.
\begin{prop}\label{prop:mpa-equiv}
If features $\Fs^\prime$ and $\Fs \setminus \Fs^\prime$ are independent given the class $C$, then $\MAA_\alpha(\Fs^\prime) = \MPA_\alpha(\Fs^\prime)$.
\end{prop}
The above property is useful because it is generally easier to compute $\MPA(\Fs^\prime)$ than $\MAA(\Fs^\prime)$, as we can maximize each instantiation $\fs^\prime$ separately.
Moreover, in naive Bayes models, the quantity MAA that we wish to optimize is now monotonic. Thus, we need to compute this quantity only for those subsets that exhaust the budget, instead of every subset that fits within budget.
Detailed proofs of above propositions can be found in the appendix.

%%%%%%%%%%%%%%%%%%%%%%%%%%%%%%%%%%%%%%%%%%%%%%%%%%%%%%%%%%%%%%%%

\section{Probabilistic Reasoning Algorithms}

In this section, we first introduce the notion of same-decision probability and formalize its connection to the ECA. We then describe our proposed algorithms to compute the MPA and the MAA that exploit this connection.

\subsection{Same-Decision Probability and ECA}
Suppose we make a decision based on whether ${\Pr(c\given \es) \geq T}$, where \(\es\) is the evidence collected thus far.
We may ask whether to collect more evidence or to commit to the current decision.
The same-decision probability (SDP) was introduced as a solution to this {\em stopping problem}~\cite{darwiche12}.
It measures how likely we are to keep our current decision even after observing more variables \(\Xs\), and is defined as follows.
\begin{defn}
Consider a Bayesian classifier \((C,\Fs,T)\) where $\Fs$ includes disjoint sets of features \(\Es\) and \(\Xs\). The \emph{same-decision probability (SDP)} for \(\Xs\) given \(\es\) is defined as
    \begin{align*}
        \SDP_{C,T}(\Xs \given \es) = \sum_{\xs} \left[C_T(\xs\es) = C_T(\es) \right] \cdot \Pr(\xs \given \es).
    \end{align*}
\end{defn}

\eat{
For example, suppose a student answers $Q_3$ in Figure~\ref{fig:nb-cpt} correctly and passes the quiz. The SDP for $\{Q_1,Q_2\}$ is 65\%, which is the probability of observing $\{Q_1\!=\!+, Q_2\!=\!+, Q_3\!=\!+\}$ or $\{Q_1\!=\!+, Q_2\!=\!-, Q_3\!=\!+\}$ given the evidence $Q_3\!=\!+$.
}
A high SDP encourages one to stop collecting information and commit to the current decision, while a low SDP suggests otherwise.
In the latter case, SDP also provides a solution to deciding which observations to collect next.
In particular, observing a subset of variables $\Ys$ in $\Xs$ that maximizes the {\em expected SDP (E-SDP)} will lead to the most robust decision {\em in expectation.} Hence, such subset maximally eliminates the need for further observations. This is formalized as follows.
\begin{defn}
Consider a Bayesian classifier \((C,\Fs,T)\) where $\Fs$ includes disjoint sets of features \(\Es\), \(\Ys\) and \(\Zs\).
The \emph{expected same-decision probability} for \(\Zs\) given \(\Ys\) and \(\es\) is defined as
    \begin{align}
        \SDP_{C,T}(\Zs \given \Ys,\es)
        = \sum_{\ys} \SDP_{C,T}(\Zs \given \ys\es) \cdot \Pr(\ys \given \es) \nonumber \\
        = \sum_{\ys\zs} \left[C_T(\ys\zs\es) = C_T(\ys\es)\right] \cdot \Pr(\ys\zs \given \es). \label{eq:esdp}
    \end{align}
\end{defn}

While the ECA and E-SDP are conceptually different notions motivated by different considerations, they are quite similar computationally as they are both expectations.
The ECA is equivalent to a variant of E-SDP (denoted $\SDP_{C,T,T^\prime}$) that uses two thresholds instead of one, by replacing the indicator term in Equation~\ref{eq:esdp} with $\left[C_T(\ys\zs\es)=C_{T^\prime}(\ys\es)\right]$.
\begin{prop}\label{prop:eca-esdp}
  Let $\alpha = (C,\Fs,T)$ be a Bayesian classifier and let $\beta = (C,\Fs^\prime,T^\prime)$ be one of its trimmings. We then have
  \begin{align*}
    \ECA(\alpha,\beta) = \SDP_{C,T,T^\prime}((\Fs \setminus \Fs^{\prime}) \given \Fs^\prime),
  \end{align*}
  where the expected SDP is computed w.r.t.~classifier \(\alpha\).
\end{prop}
Therefore, one can utilize an E-SDP algorithm to compute ECA during classifier trimming.
For example, the E-SDP algorithm by~\citet{ChoiIJCAI17} can be modified to evaluate the above variant of E-SDP without extra computational overhead.

\subsection{Computing the MPA}

We now describe how we compute the MPA at each search step. First, the MPA can be expressed as follows:
\begin{align}
\MPA(\Fs^\prime) = 
&\sum_{\fs^\prime} \max \Big(\SDP_{C,T,0}(\Rs\vert\fs^\prime), \nonumber \\
&\phantom{\sum_{\fs^\prime} \max \Big(} 1-\SDP_{C,T,0}(\Rs\vert\fs^\prime)\Big)\cdot\Pr(\fs^\prime). \label{eq:mpa-sdp}
\end{align}
Here, $\SDP_{C,T,0}(\Rs\vert\fs^\prime)$ is the expected probability that we will decide positive class if we observe features $\Rs$ given $\fs^\prime$. Then $1 - \SDP_{C,T,0}(\Rs\vert\fs^\prime)$ is its complement: the expected probability of negative classification.
Exploiting this connection to SDP, our algorithm makes heavy use of the E-SDP algorithm by~\citet{ChoiIJCAI17}, of which we provide a high-level description here and refer to the original paper for details.
The algorithm is based on compiling a Bayesian network into a tractable circuit representation, called a Sentential Decision Diagram (SDD)~\cite{Darwiche11}. Even though compiling the circuit is computationally heavy in general, computing the E-SDP and hence the MPA is efficient once we have successfully compiled the circuit.
Moreover, we can sometimes efficiently compile certain networks (e.g. high treewidth) in which traditional inference techniques become infeasible~\cite{ChoiKisaDarwiche13}.
As a part of the process to compute $\SDP_{C,T}(\Rs\given\Fs^\prime)$, the E-SDP algorithm calculates and saves the values $\Pr(\fs^\prime)$ and $\SDP_{C,T,0}(\Rs\given\fs^\prime)$ for each $\fs^\prime$. Given these values, computing the MPA for $\Fs^\prime$ is a straightforward evaluation of Equation~\ref{eq:mpa-sdp}.

With the ability to compute the MPA, we can now search for optimal trimmings of naive Bayes classifiers.
The condition in Proposition~\ref{prop:mpa-equiv} holds for all feature subsets of a naive Bayes model, and thus the MAA of a subset is always equal to its MPA.
An optimal trimming is then found as shown in Algorithm~\ref{alg:trim-nb} where both the upper bound and value of MAA are computed using the MPA algorithm described before.

\begin{algorithm}[tb]
\caption{$\textsc{Compute-MAA}$}\label{alg:maa}
\textbf{Input:} \\
  $\alpha$ : Bayesian network classifier $(C,\Fs,T)$;\quad $\Fs^\prime \subset \Fs$ \\
\textbf{Data:} \\
  $\text{CPR}(i) \leftarrow \Pr(c \given \fs_i^\prime)$ for all $i$ \\
  $\text{MAR}(i) \leftarrow \Pr(\fs_i^\prime)$ for all $i$ \\
  $\text{POS}(i) \leftarrow \SDP_{C,T,0}((\Fs\setminus\Fs^\prime)\given\fs_i^\prime)$ for all $i$ \\
\textbf{Output:} The score $\MAA(\Fs^\prime)$ and the optimal threshold $T^\prime$
\\\hrule
  \begin{algorithmic}[1]
    \State Sort instances $\fs_i^\prime$ in nondecreasing order of $\text{CPR}(i)$
    \State $m \leftarrow \sum_i \text{POS}(i)\cdot\text{MAR}(i)$;\quad$m^\star \leftarrow m$
    \State $t^\star \leftarrow [0,\text{CPR}(1)]$
    \For {$i$ in $1,2,\dots$}
        \State $m \leftarrow m - \text{MAR}(i)\cdot(2\text{POS(i)} - 1)$ \label{line:eca-update}
        \If {$m > m^\star$} 
            \State $m^\star \leftarrow m$;\;$t^\star \leftarrow (\text{CPR}(i), \text{CPR}(i+1)]$
        \EndIf
    \EndFor
    \State \textbf{return} $\MAA(\Fs^\prime)=m^\star$ and any $T^\prime \in t^\star$
  \end{algorithmic}
\end{algorithm}

\subsection{Computing the MAA}

Searching for an optimal trimming of arbitrary Bayesian network classifiers requires the computation of MAA, which involves tuning the trimmed classifier's threshold to maximize the ECA.
First, we utilize the observation that a change in threshold affects the value of ECA only if the class probability given some instance lies on a different side of the threshold after the change.
For example, recall the trimmed classifier using only $Q_3$ from the quiz example in Section~\ref{sec:eca}. We showed that a threshold of $0.15$ will result in passing only the students who answered $Q_3$ correctly.
In fact, any threshold between $\Pr(C\!=\!+ \vert Q_3\!=\!-)\!=\!0.08$ and $\Pr(C\!=\!+ \vert Q_3\!=\!+)\!=\!0.18$ results in the same behavior and hence the same ECA.
Therefore, the number of threshold values we need to consider is finite and in fact linear in the number of possible instances $\fs^\prime$.

\eat{
One way to calculate $\MAA(\Fs^\prime)$ is to compute the ECA for every candidate threshold $T^\prime$ and choose the maximum value, where the value of ECA is obtained using the E-SDP algorithm, exploiting the connection shown in Proposition~\ref{prop:eca-esdp}.
However, not only is repeated execution of the algorithm very expensive, but it will also result in many redundant calculations.
Consider for example the ECA computations for two trimmed classifiers $\beta_1=(C,\Fs^\prime,T_1)$ and $\beta_2=(C,\Fs^\prime,T_2)$ with respect to the same original classifier, with $T_1 \leq T_2$.
Then their classification behaviors only differ in those observations $\fs^\prime$ that lead to a class probability $\Pr(c\given\fs^\prime)$ between $T_1$ and $T_2$.
Hence, any computation regarding the other observations become redundant if the algorithm is run twice for $\beta_1$ and $\beta_2$.
}

To compute the MAA using this observation, we first express the ECA as the following:
\begin{align}
\ECA(\alpha,\beta) = 
\sum_{\fs^\prime:\Pr(c|\fs^\prime)\geq T^\prime} \SDP_{C,T,0}(\Rs\given\fs^\prime)\cdot\Pr(\fs^\prime) \nonumber \\
\phantom{=} + \sum_{\fs^\prime:\Pr(c|\fs^\prime)< T^\prime} \Big(1 -\SDP_{C,T,0}(\Rs\given\fs^\prime)\Big)\cdot\Pr(\fs^\prime). \label{eq:eca-alt}  
\end{align}
Then we can compute the ECA for any given threshold $T^\prime$ if we have the values of $\Pr(c\given\fs^\prime)$, $\Pr(\fs^\prime)$, and $\SDP_{C,T,0}(\Rs\given\fs^\prime)$ for each instance $\fs^\prime$, which we can indeed easily obtain from an execution of the E-SDP algorithm for $\SDP_{C,T}(\Rs\given\Fs^\prime)$.

Algorithm~\ref{alg:maa} shows the pseudocode to compute the MAA and the optimal threshold given these values from the E-SDP algorithm as inputs.
Starting from $T^\prime = 0$, the threshold is repeatedly incremented to just above the next lowest class probability given some feature instance $\fs^\prime$.
With each threshold change, the ECA value is also updated by subtracting the expected probability of positive classification given that instance and adding the complement of it, weighted by the marginal probability of that instance, as in Line~\ref{line:eca-update}.
In other words, that instance $\fs^\prime$ is moved from the first sum to the second in Equation~\ref{eq:eca-alt}.
At the end, the highest value of ECA and its corresponding threshold is reported.

\begin{figure}[tb]
  \centering
  % \scalebox{0.9}{
  \begin{tikzpicture}[bayesnet]
      
  \def\lone{0bp}
  \def\ltwo{-35bp}

  \node (F1) at (-65bp,\lone) [bnnode] {$F_1$};
  \node (C) at (-25bp,\lone) [bnnode] {$C$};
  \node (F2) at (-65bp,\ltwo) [bnnode] {$F_2$};
  \node (F3) at (-25bp,\ltwo) [bnnode] {$F_3$};
  
  \begin{scope}[on background layer]
    \draw [bnarrow] (C) -- (F2);
    \draw [bnarrow] (C) -- (F3);
    \draw [bnarrow] (F1) -- (F2);
    \draw [bnarrow] (F2) -- (F3);
  \end{scope}
  
  \node at (37bp,0bp) [cpt, anchor=north] {
    \scriptsize
    \begin{tabular}{c}
      \toprule
      $\Pr(F_1\!=\!+)$\\\midrule
      $0.9$\\
      \bottomrule
    \end{tabular}
  };

  \node at (92bp,0bp) [cpt, anchor=north] {
    \scriptsize
    \begin{tabular}{c}
      \toprule
      $\Pr(C\!=\!+)$\\\midrule
      $0.6$\\
      \bottomrule
    \end{tabular}
  };
    
  \node at (-45bp,-45bp) [cpt, anchor=north] {
    \scriptsize
    \begin{tabular}{llc}
      \toprule
      $C$ & $F_1$ & $\Pr(F_2\!=\!+|C F_1)$\\\midrule
      $+$ & $+$ & $0.6$\\
      $+$ & $-$ & $1.0$\\
      $-$ & $+$ & $0.4$\\
      $-$ & $-$ & $0.5$\\
      \bottomrule
    \end{tabular}
  };
  
  \node at (67bp,-45bp) [cpt, anchor=north] {
    \scriptsize
    \begin{tabular}{llc}
      \toprule
      $C$ & $F_2$ & $\Pr(F_3\!=\!+|C F_2)$\\\midrule
      $+$ & $+$ & $0.4$\\
      $+$ & $-$ & $1.0$\\
      $-$ & $+$ & $1.0$\\
      $-$ & $-$ & $0.4$\\
      \bottomrule
    \end{tabular}
  };
  \end{tikzpicture}
  % }
  \caption{A Bayes net over features $\{F_1,F_2,F_3\}$ and class~$C$.}\label{fig:gbn-cpt}
\end{figure}
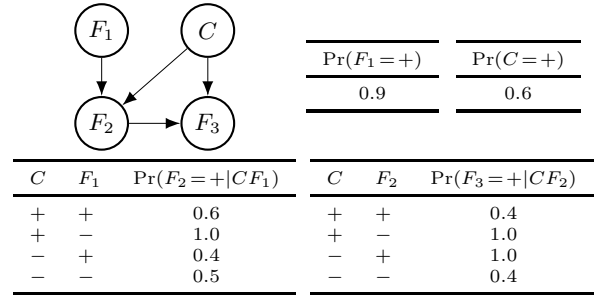

\begin{table}[tb]
  \centering
  \scalebox{0.9}{
  \begin{tabular} { c c c c }
    \toprule
    $\{F_1,F_2\}$ & $\Pr(c \given F_1 F_2)$ & $+$ class pr. & $-$ class pr. \\
    \midrule
    $\{-,+\}$   & 0.75  & 0.04  & 0.04 \\
    $\{+,+\}$   & 0.69  & 0.20  & 0.27 \\
    $\{+,-\}$   & 0.50  & 0.30  & 0.13 \\
    \hdashline
    $\{-,-\}$   & 0.00  & 0.00  & 0.02 \\
    \bottomrule
  \end{tabular}}
  \caption{Table to calculate the $\MAA(\{F_1,F_2\})$}\label{table:eca}
\end{table}

Table~\ref{table:eca} offers a visualization of the algorithm.
Here, we wish to compute $\MAA(\{F_1,F_2\})$ with respect to the Bayesian network classifier $\alpha = (C,\{F_1,F_2,F_3\},0.55)$ in Figure~\ref{fig:gbn-cpt}.
Each table row corresponds to a feature instance, sorted by the class probability. 
We consider five different cutoff points, and the ECA value at each cutoff point is the sum of expected probability of positive class for instances above the line and the expected probability of negative class below the line.
In this case, $\MAA(\{F_1,F_2\}) = 0.56$ with the optimal threshold $T^\star \in (0,0.50]$, indicated by a dotted line in the table.

%%%%%%%%%%%%%%%%%%%%%%%%%%%%%%%%%%%%%%%%%%%%%%%%%%%%%%%%%%%%%%%%

\begin{figure*}[tb]
  \centering
  \subcaptionbox{ECA and average accuracy for \texttt{pima}} {    
    \centering
    \scalebox{0.8}{    
    \begin{tikzpicture}
      \begin{axis}[
          title={},
          width=7cm, height=5cm,
          xlabel={Average Accuracy},
          ylabel={ECA},
          y label style={at={(axis description cs:0.1,.5)},anchor=south},
          legend pos=north west,
          legend style={nodes={scale=0.9}},          
          legend entries={Feasible $\Fs'$, Optimal ECA, Optimal Accuracy},
          scatter/classes={
          X={mark=x,mark size=1pt},%
          E={mark=square*,blue,mark size=2pt},%
          A={mark=*,black!30!green,mark size=2pt}}]
        \addplot[scatter,only marks, mark size=1pt,scatter src=explicit symbolic]
        table[x=ACC,y=ECA,meta=Type,col sep=comma]{pima-eca-acc.csv};
      \end{axis}
    \end{tikzpicture}
    }
  }
  \subcaptionbox{ECA and average accuracy for \texttt{heart}} {
    \centering
    \scalebox{0.8}{    
    \begin{tikzpicture}
      \begin{axis}[
          title={},
          width=7cm, height=5cm,
          xlabel={Average Accuracy},
          ylabel={ECA},
          y label style={at={(axis description cs:0.1,.5)},anchor=south},
          legend pos=north west,
          legend style={nodes={scale=0.9}},
          legend entries={Feasible $\Fs'$, Optimal ECA, Optimal Accuracy},          
          scatter/classes={
          X={mark=x,mark size=1pt},%
          E={mark=square*,blue,mark size=2pt},%
          A={mark=*,black!30!green,mark size=2pt}}]
        \addplot[scatter,only marks, mark size=0.7pt,scatter src=explicit symbolic]
        table[x=ACC,y=ECA,meta=Type,col sep=comma]{heart-eca-acc.csv};
      \end{axis}
    \end{tikzpicture}
    }
  }
  \subcaptionbox{Test agreement and accuracy\label{fig:test-eca-acc}} {
    \centering
    \raisebox{30mm}{
    \scalebox{0.8}{
    \begin{tabular}[t]{llcc}
      \toprule
      & & Agreement & Accuracy \\
      \midrule
      \multirow{2}{*}{\texttt{pima}} & Opt. ECA & 0.9863 & 0.7123 \\
      & Opt. Acc. & 0.9452 & 0.7260 \\
      \midrule
      \multirow{2}{*}{\texttt{heart}} & Opt. ECA & 0.9245 & 0.8491 \\
      & Opt. Acc. & 0.9057 & 0.7925 \\
      \bottomrule
    \end{tabular}
    }
    }
  }
  \caption{(a),(b) ECA and average accuracy achieved by feasible feature subsets. (c) evaluation of subsets with highest ECA and accuracy.}\label{fig:eca-accuracy}
\end{figure*}
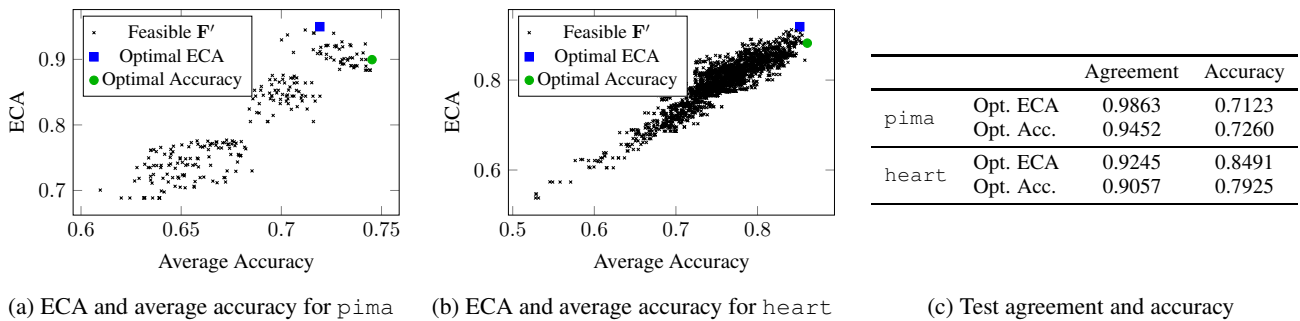

\subsection{Complexity}

Our proposed MPA and MAA algorithms, as described so far, appear to consider all possible feature instances, which is exponential in the size of the feature subset.
Nevertheless, this can be improved by exploiting context-specific independence, a property having to do with the parameters of the classifier's probability distribution~\cite{boutilier1996context}.
In particular, suppose given a context $\gs$ (where $\Gs \subseteq \Fs^\prime$), the remaining features $\Fs^\prime \setminus \Gs$ become independent of the class variable. Then in our MPA and MAA calculations, we only need to consider the context $\gs$ instead of individually considering all instances of $\Fs^\prime$ that share this partial instantiation.
The E-SDP algorithm by~\citet{ChoiIJCAI17} that we utilize is based on SDD compilation, which exploits some context-specific independence to simplify the circuit.\footnote{We refer to~\citet{chavira2008probabilistic} and~\citet{ChoiKisaDarwiche13} for more details on this simplification.}
However, this is currently limited to certain cases, such as when some features have uniform parameters given a context, and does not exploit all simplification opportunities described above.
Hence, compilation that fully exploits context-specific independence is an interesting future direction to further improve the efficiency of our algorithms.

\section{Experimental Evaluation}
We now empirically evaluate our proposed algorithms on several naive Bayes and general Bayesian network benchmarks.

\subsection{Accuracy vs.~Agreement}

We evaluate our method on real-world datasets from the UCI repository \cite{BacheLichman2013}, BFC (\url{http://www.berkeleyfreeclinic.org/}), and CRESST (\url{http://www.cse.ucla.edu/}).
We randomly split each dataset into 80/20 train and test sets and learn a naive Bayes classifier using the training set. With the budget set as half the number of features and threshold as 0.5, we compute the ECA of each feasible feature subset.
In addition, we compute the average classification accuracy of each feature subset using 10-fold cross validation on the training set.

Figure~\ref{fig:eca-accuracy} shows the ECA and average accuracy achieved by each feasible subset, and the subsets with highest ECA and highest accuracy are highlighted.
We can see that optimizing the ECA tends towards higher accuracy. More interestingly, we can observe a Pareto frontier where one cannot increase the ECA without sacrificing average accuracy, and vice versa.
This suggests that one may need to make a tradeoff between classifier agreement (i.e., robustness) and accuracy when selecting features.
Moreover, we evaluate the subsets with highest ECA and accuracy on the test set and report their empirical classification agreement and accuracy in Figure~\ref{fig:test-eca-acc}. 
The subset chosen for optimal ECA on the training set also achieves high classification agreement on the test set.
Surprisingly, on network \texttt{heart}, it also achieves higher test accuracy than the subset with the highest average cross-validation accuracy on the training set.
A possible explanation is that choosing subsets based on their cross-validation classification accuracy does not generalize well to the test set. It may introduce additional overfitting that ECA does not suffer from: if the original model generalizes well, we also expect our trimmed classifier to generalize well.
We also evaluated accuracy and agreement of feature selection by information gain, but it neither outperformed optimizing the ECA nor the average cross-validation accuracy.
In addition, our method achieves higher accuracy on most splits of the \texttt{heart} data, which suggests that this may be a property of the dataset.
In particular, the average cross-validation accuracy of the original full classifier for \texttt{pima} was approximately 0.720, which was lower than the average accuracy of about 21\% of the candidate subsets.
As our method optimizes for agreement with this original classifier, which has relatively low accuracy, the resulting trimming may have lower accuracy than if we were to actively optimize for average accuracy.
On the other hand, the original classifier for \texttt{heart} had average accuracy 0.866, which was lower than only 2\% of the candidate subsets.
Hence, in this case, optimizing the ECA to closely mimic the original classifier's behavior also results in relatively high classification accuracy.

\begin{table}[tb]
  \centering
  \scalebox{0.8}{
  \begin{tabular}{cc*{5}{r}}
    \toprule
    & & \multicolumn{2}{c}{FS-SDD} & \multicolumn{2}{c}{\textsc{ECA-Trim}} \\
    & \multicolumn{1}{c}{$|\Fs|$}& \multicolumn{1}{c}{Time} & \multicolumn{1}{c}{\# Eval} & \multicolumn{1}{c}{Time} & \multicolumn{1}{c}{\# Eval} & \multicolumn{1}{c}{$\binom{n}{m}$} \\
    \midrule
    \texttt{bupa}    & 6  & 0.044    & 21   & 0.026  & 14  & 15 \\
    \texttt{pima}    & 8  & 0.056    & 36   & 0.039  & 45  & 28 \\
    \texttt{ident}   & 9  & 0.128    & 129  & 0.097  & 89  & 84 \\
    \texttt{anatomy} & 12 & 2.252    & 793  & 1.085  & 283 & 495 \\
    \texttt{heart}   & 13 & 7.346    & 1092 & 2.234  & 209 & 715 \\
    \texttt{voting}  & 16 & 819.163  & 6884 & 407.571& 3345& 4368 \\
    \texttt{hepatitis} &19& Timeout  & 43795& 4390.71& 2208& 27132 \\
    \bottomrule
  \end{tabular}}
  \caption{Runtime in seconds and number of criteria evaluations}\label{table:nb-comp}
\end{table}

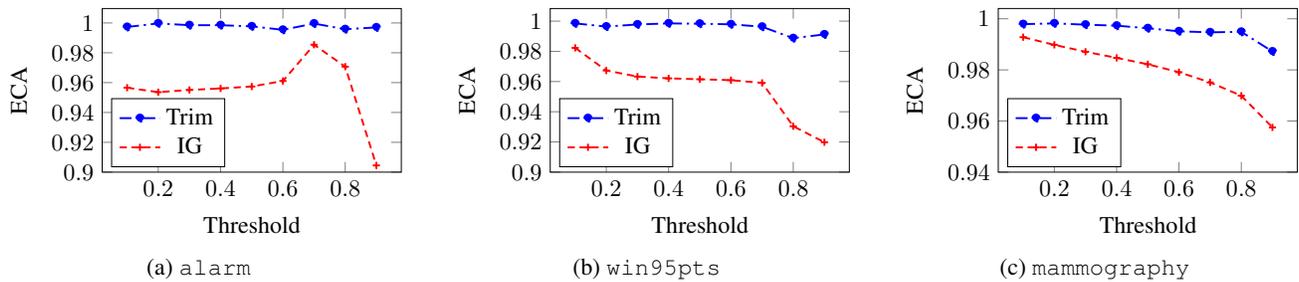
\begin{figure*}[tb]
\centering
\subcaptionbox{\texttt{alarm}}[0.33\textwidth] {
  \centering
  \scalebox{0.9}{
  \begin{tikzpicture}
    \begin{axis}[
      title={},
      xlabel={Threshold},
      ylabel={ECA},
      ymin=0.90,
      ytick={0.90,0.92,0.94,0.96,0.98,1.00},
      legend pos=south west,
      width=6cm, height=4cm,
    ]
    \addplot[
      thick,
      dash pattern={on 5pt off 2pt on 1pt off 2pt},
      color=blue,
      mark=*,
      mark size = 1.5pt
      ]
      coordinates {
      (0.1,0.9973)(0.2,0.9997)(0.3,0.9985)(0.4,0.9985)(0.5,0.9977)(0.6,0.9954)(0.7,0.9996)(0.8,0.9958)(0.9,0.9970)
      };
      \addlegendentry{Trim}
    \addplot[
      thick,
      densely dashed,
      color=red,
      mark=+,
      mark size = 1.5pt
      ]
      coordinates {
      (0.1,0.9566)(0.2,0.9536)(0.3,0.9551)(0.4,0.9561)(0.5,0.9574)(0.6,0.9611)(0.7,0.9854)(0.8,0.9707)(0.9,0.9045)
      };
      \addlegendentry{IG}
    \end{axis}
  \end{tikzpicture}
  }
}
\subcaptionbox{\texttt{win95pts}}[0.33\textwidth] {
  \centering
  \scalebox{0.9}{
  \begin{tikzpicture}
    \begin{axis}[
      title={},
      xlabel={Threshold},
      ylabel={ECA},
      ymin=0.90,
      ytick={0.90,0.92,0.94,0.96,0.98,1.00},
      legend pos=south west,
      width=6cm, height=4cm,
    ]
    \addplot[
      thick,
      dash pattern={on 5pt off 2pt on 1pt off 2pt},
      color=blue,
      mark=*,
      mark size = 1.5pt
      ]
      coordinates {
      (0.1,0.9985)(0.2,0.9964)(0.3,0.9978)(0.4,0.9985)(0.5,0.9983)(0.6,0.9979)(0.7,0.9963)(0.8,0.9887)(0.9,0.9912)
      };
      \addlegendentry{Trim}
    \addplot[
      thick,
      densely dashed,
      color=red,
      mark=+,
      mark size = 1.5pt
      ]
      coordinates {
      (0.1,0.9823)(0.2,0.9673)(0.3,0.9633)(0.4,0.9621)(0.5,0.9615)(0.6,0.9609)(0.7,0.9591)(0.8,0.9304)(0.9,0.9198)
      };
      \addlegendentry{IG}
    \end{axis}
  \end{tikzpicture}
  }
}
\subcaptionbox{\texttt{mammography}}[0.33\textwidth] {
  \centering
  \scalebox{0.9}{
  \begin{tikzpicture}
    \begin{axis}[
      title={},
      xlabel={Threshold},
      ylabel={ECA},
      ymin=0.94,
      ytick={0.94,0.96,0.98,1.00},
      legend pos=south west,
      width=6cm, height=4cm,
    ]
    \addplot[
      thick,
      dash pattern={on 5pt off 2pt on 1pt off 2pt},
      color=blue,
      mark=*,
      mark size = 1.5pt
      ]
      coordinates {
      (0.1,0.9979)(0.2,0.9982)(0.3,0.9977)(0.4,0.9973)(0.5,0.9963)(0.6,0.9951)(0.7,0.9947)(0.8,0.9949)(0.9,0.9872)
      };
      \addlegendentry{Trim}
      \addplot[
      thick,
      densely dashed,
      color=red,
      mark=+,
      mark size = 1.5pt
      ]
      coordinates {
      (0.1,0.9928)(0.2,0.9898)(0.3,0.9871)(0.4,0.9847)(0.5,0.9822)(0.6,0.9791)(0.7,0.9751)(0.8,0.9699)(0.9,0.9575)
      };
      \addlegendentry{IG}
    \end{axis}
  \end{tikzpicture}
  }
}
\caption{Comparing ECA of features selected by classifier trimming and information gain}\label{fig:eca-ig}
\end{figure*}

\subsection{Runtime}
We also compare the efficiency of our algorithm against FS-SDD, which is the feature selection algorithm based on E-SDP~\cite{ChoiIJCAI17}.
Each naive Bayes classifier was trimmed with the budget set to \nicefrac{1}{3} the number of features, each feature given unit cost, and classification thresholds in $\{0.1,0.2,\dots,0.9\}$.
\eat{All experiments were performed with a 2.6GHz Intel Xeon E5-2670 CPU with a memory limit of 4GB RAM.}
Table~\ref{table:nb-comp} shows the average running times in seconds and the number of times each algorithm computes the criterion value. 
The performance of our method is comparable to that of FS-SDD on smaller classifiers, and more efficient for larger ones.
In particular, our algorithm could handle the largest network with around 4000 ECA computations, whereas an exhaustive search method by FS-SDD requires a significantly greater number of computations and could not be run to completion.
Moreover, in 16\% of experiment instances described above, the trimmed classifier with optimized threshold reported higher classification similarity (measured by the ECA) than the classifier using features selected by FS-SDD, which keeps the threshold fixed.
Note that our method optimizes the threshold of the trimmed classifier and thus is theoretically guaranteed to achieve classification similarity (measured by the ECA) no lower than feature selection using E-SDP. Hence, threshold optimization allows us to find more robust trimmings more efficiently.
We also show the total number of subsets within budget, $\binom{n}{m}$, to illustrate that branch-and-bound algorithms tend to require fewer number of evaluations than exhaustive search as the number of features increase.
This justifies the extra MPA calculations done for pruning on subsets larger than the budget.

\subsection{Trimming General Networks}
Next, we evaluate the quality of trimmed classifiers on general Bayesian networks from the UAI 2008 evaluation and a tree-augmented naive Bayes model for mammography reports~\cite{gimenez2014novel}.
We run our method with $T$ in $\{0.1,0.2,\dots,0.9\}$ and the budget set to \nicefrac{1}{3} the number of features.
For the UAI networks, we randomly chose a root node to be the class variable and used the set of all leaf nodes as the feature set \Fs. Each setting was repeated for three randomly selected class variables.
For the mammography network, we used the (root) decision node as the class variable and chose 17 out of the 20 variables to be the feature set.
Training data was not available for these networks, so we compare against feature selection by information gain instead of classification accuracy. Figure~\ref{fig:eca-ig} highlights the results.
The trimmed classifier by our algorithm consistently achieves higher expected classification agreement, demonstrating that robustness is not easily achieved by other feature selection methods.
\eat{For network ‘alarm’, the average ECA of trimmed classifiers by our algorithm is 99.8\%, whereas it is 95.6\% using information gain. For network ‘win95pts’, the average ECA values are 99.6\% versus 95.6\%. In some settings, the differences in ECA values were as large as 28\%.
Therefore, our feature trimming algorithm returns a classifier that is highly similar to the original classifier, a criteria that is not easily reached by other feature selection algorithms.}
We also want to stress that the features selected using information gain differ for different class variables, but stay the same across different initial threshold values.
On the other hand, our algorithm is sensitive to the original threshold, and thus results in trimmed classifiers with similar behavior as the original classifier with a particular threshold.

\section{Conclusion}

This paper developed a novel operator on Bayesian classifiers: to trim the set of features to fit within a budget, while simultaneously adjusting the classification threshold. Our objective was to optimize the expected classification agreement between the original classifier and its trimmed counterpart. By analyzing the properties of classifier agreement and its maximum potential agreement, we developed a branch-and-bound search algorithm to find optimal trimmings.
Experiments on naive and general Bayesian networks demonstrated the effectiveness of our approach in finding robust trimmings of classifiers, especially compared to optimizing more traditional objectives such as expected SDP and information gain.

\section*{Acknowledgments}
The authors wish to thank Adnan Darwiche for his contributions to an earlier version of this work, and Arthur Choi for helpful advice and discussions.
This work is partially supported by NSF grants \#IIS-1657613, \#IIS-1633857 and DARPA XAI grant \#N66001-17-2-4032.

% \FloatBarrier

\bibliographystyle{named}
\bibliography{references}

% \eat{
\appendix

\section{Proof of Proposition~\ref{prop:mpa}}
\begin{proof}
Let $\Rs = \Fs \setminus \Fs^\prime$. The proposition follows from the definitions of MAA and MPA as follows:
\begin{align}
    &\MAA_{\alpha}(\Fs^\prime)
    = \max_{T^\prime} \sum_{\fs^\prime \rs} \left[C_T(\fs^\prime \rs) = C_{T^\prime}(\fs^\prime)\right] \cdot \Pr(\fs^\prime \rs) \nonumber\\
    &\leq \sum_{\fs^\prime} \max_{T^\prime} \sum_\rs \left[C_T(\fs^\prime \rs) = C_{T^\prime}(\fs^\prime)\right] \cdot \Pr(\fs^\prime \rs) \nonumber\\
    &= \sum_{\fs^\prime} \max_c \left\{\sum_{\rs} \left[C_T(\fs^\prime\rs)=c\right]\cdot\Pr(\fs^\prime\rs) \right\} \label{eq:mpa-pf-c}\\
    &= \MPA_\alpha(\Fs^\prime). \nonumber
\end{align}
Equation~\ref{eq:mpa-pf-c} comes from the fact that, for a fixed instance $\fs^\prime$, choosing a threshold $T^\prime$ is equivalent to choosing to classify that instance positively or negatively.
\end{proof}

\section{Proof of Proposition~\ref{prop:mpa-mono}}
\begin{proof}
Let $\Rs_1 = \Fs \setminus \Fs_1$ and $\Rs_2 = \Fs \setminus \Fs_2$.
Say $\Gs = {\Fs_2 \setminus \Fs_1} = {\Rs_1 \setminus \Rs_2}$. Then,
\begin{align}
  \MPA(\Fs_1) 
  &= \sum_{\fs_1} \max_c \left\{\sum_{\gs\rs_2} \left[C_T(\fs_1\gs\rs_2)=c\right]\cdot\Pr(\fs_1\gs\rs_2) \right\} \label{eq:mpa-pf-sub} \\
  &\leq \sum_{\fs_1\gs} \max_c \left\{\sum_{\rs_2} \left[C_T(\fs_1\gs\rs_2)=c\right]\cdot\Pr(\fs_1\gs\rs_2) \right\}
  \label{eq:mpa-pf-ineq} \\
  &= \MPA(\Fs_2). \nonumber
\end{align}
Equation~\ref{eq:mpa-pf-sub} is from the assumption that $\Rs_1 = \Gs \cup \Rs_2$, and Equation~\ref{eq:mpa-pf-ineq} follows from the fact that sum of maxima is an upper bound on the maximum of sums.
\end{proof}

\section{Proof of Proposition~\ref{prop:mpa-equiv}}
\begin{proof}
We have $\MAA(\Fs^\prime) \leq \MPA(\Fs^\prime)$ from Proposition~\ref{prop:mpa}. To show $\MAA(\Fs^\prime) \geq \MPA(\Fs^\prime)$ under the independence assumption, we will use the following claim.
\begin{claim}
  Suppose features $\Fs^\prime$ and $\Rs$ are independent given class $C$. For any instances $\fs^\prime_1$ and $\fs^\prime_2$, if $\Pr(c \given \fs^\prime_1) \geq \Pr(c \given \fs^\prime_2)$, then at least one of the following must hold:
  \begin{enumerate}
      \item $\sum_{\rs} \left[C_T(\fs^\prime_1\rs)=c\right]\cdot\Pr(\rs\vert\fs^\prime_1) \geq 0.5$, or \label{enum:claim1}
      \item $\sum_{\rs} \left[C_T(\fs^\prime_2\rs)=\overline{c}\right]\cdot\Pr(\rs\vert\fs^\prime_2) \geq 0.5$. \label{enum:claim2}
  \end{enumerate}
  If inequality~\ref{enum:claim1} is true, we say $\fs^\prime_1$ ``favors positive classification'', and if inequality~\ref{enum:claim2} is true, we say $\fs^\prime_2$ ``favors negative classification''.
\end{claim}
Above claim states that, if $\Fs^\prime$ and $\Rs$ are independent given $C$ and if positive class $c$ is more likely given $\fs^\prime_1$ than given $\fs^\prime_2$, then we cannot have $\fs^\prime_1$ favor negative classification while $\fs^\prime_2$ favor positive classification.

Assuming the claim is true, we can choose a $T^\prime$ such that $Q_1 < T^\prime \leq Q_2$, where $Q_1 = \max_{\fs^\prime} \Pr(c \given \fs^\prime)$ such that $\fs^\prime$ favors negative class and $Q_2 = \min_{\fs^\prime} \Pr(c \given \fs^\prime)$ such that $\fs^\prime$ favors positive class.
Then, an instance $\fs^\prime$ favors positive class if and only if $\Pr(c \given \fs^\prime) \geq T^\prime$. Thus, we obtain the following inequality:
\begin{align}
  &\MAA(\Fs^\prime)
  \geq \sum_{\fs} \left[C_T(\fs) = C_{T^\prime}(\fs^\prime)\right]\cdot\Pr(\fs) \label{maa-pf-ineq} \\
  &= \sum_{\fs^\prime} \max_c \left\{\sum_{\rs} \left[C_T(\fs^\prime\rs)=c\right]\cdot\Pr(\rs \given \fs^\prime) \right\} \cdot \Pr(\fs^\prime) \label{maa-pr-max} \\
  & = \MPA(\Fs^\prime) \nonumber
\end{align}

Equation~\ref{maa-pf-ineq} follows from the definition of MAA as the maximum of ECA across different $T^\prime$.
Equation~\ref{maa-pr-max} holds because $T^\prime$ was explicitly constructed such that the trimmed classifier classifies each instance $\fs^\prime$ into the class that it favors in the original classifier.

Now we prove the claim by considering two possible cases.
Suppose there exists $\fs^\prime_1$ and $\fs^\prime_2$ such that $\Pr(c \given \fs^\prime_1) \geq \Pr(c \given \fs^\prime_2)$ but $\fs^\prime_1$ favors negative class while $\fs^\prime_2$ favors positive class. Since $\Fs^\prime$ and $\Rs$ are independent given $C$, we can write the following in log-odds domain: $\log O(c \given \fs^\prime\rs) = \log O(c) + w_{\fs^\prime} + w_{\rs}$, where $w_\xs=\log \frac{\Pr(\xs \given c)}{\Pr(\xs \given \overline{c})}$.
Then $\Pr(c \given \fs^\prime\rs) \geq T$ if and only if $\log O(c \given \fs^\prime\rs) \geq \lambda = \log \frac{T}{1-T}$.

\paragraph{Case 1:} $\Pr(c \given \fs^\prime_2) < T$.
Equivalently, $\log O(c \given \fs^\prime_2) < \lambda$. Also, $\log O(c \given \fs^\prime_2) \leq \log O(c \given \fs^\prime_1)$ by assumption. Then $\forall\,\rs\: \log O(c \given \fs^\prime_2\rs) \geq \lambda \implies \log O(c \given \fs^\prime_1\rs) \geq \lambda$. For such $\rs$, $w_{\rs} > 0$ and thus $\Pr(\rs \given c) > \Pr(\rs \given \overline{c})$, which implies:
\begin{align*}
  \Pr(\rs \given \fs^\prime_1)
  &= \Pr(\rs \given c)\Pr(c \given \fs^\prime_1) + \Pr(\rs \given \overline{c})\Pr(\overline{c} \given \fs^\prime_1) \\
  &= \Pr(\rs \given \fs^\prime_2) + \alpha(\Pr(\rs \given c) - \Pr(\rs \given \overline{c})) \\
  &\geq \Pr(\rs \given \fs^\prime_2),
\end{align*}
where $\alpha = \Pr(c \given \fs^\prime_1) - \Pr(c \given \fs^\prime_2) \geq 0$.\\
Combining these, we have
\begin{align*}
  &\sum_{\rs} \left[C_T(\fs^\prime_1\rs)=c\right]\Pr(\rs \given \fs^\prime_1) \\
  &\geq \sum_{\rs} \left[C_T(\fs^\prime_2\rs)=c\right]\Pr(\rs \given \fs^\prime_2)
  > 0.5,
\end{align*}
which is a contradiction of our assumption that $\fs_1^\prime$ favors negative class (i.e. $\sum_{\rs} \left[C_T(\fs^\prime_1\rs)=c\right]\Pr(\rs \given \fs^\prime_1) < 0.5$).

\paragraph{Case 2:} $\Pr(c \given \fs^\prime_2) \geq T$.
Similarly, $\forall\,\rs\:\Pr(c \given \fs^\prime_1\rs) < T$ implies $\Pr(c \given \fs^\prime_2\rs) < T$ and $\Pr(\rs \given c) < \Pr(\rs \given \overline{c})$. Thus,
\begin{align*}
  \Pr(\rs \given \fs^\prime_2)
  &= \Pr(\rs \given c)\Pr(c \given \fs^\prime_2) + \Pr(\rs \given \overline{c})\Pr(\overline{c} \given \fs^\prime_2) \\
  &= \Pr(\rs \given \fs^\prime_1) + \alpha(\Pr(\rs \given \overline{c}) - \Pr(\rs \given c)) \\
  &\geq \Pr(\rs \given \fs^\prime_1).
\end{align*}
This leads to the following:
\begin{align*}
  &\sum_{\rs} \left[C_T(\fs^\prime_2\rs)=\overline{c}\right]\Pr(\rs \given \fs^\prime_2) \\
  &\geq \sum_{\rs} \left[C_T(\fs^\prime_1\rs)=\overline{c}\right]\Pr(\rs \given \fs^\prime_1)
  > 0.5,
\end{align*}
which is a contradiction of our assumption that $\fs_2^\prime$ favors positive class and thus $\sum_{\rs} \left[C_T(\fs^\prime_2\rs)=\overline{c}\right]\Pr(\rs \given \fs^\prime_2) < 0.5$.
This concludes the proof of the claim and the proposition.
\end{proof}

\section{Relationship to SDP: Details}

Here we show the details of how the MPA and ECA can be expressed in relation to the SDP.
First, Proposition~\ref{prop:eca-esdp} simply follows from the definition of E-SDP and ECA:
\begin{align*}
  \SDP_{C,T,T^\prime}((\Fs \setminus \Fs^{\prime}) \given \Fs^\prime) 
  &= \sum_{\fs} \left[C_T(\fs) = C_{T^\prime}(\fs^\prime) \right] \cdot \Pr(\fs) \\
  &= \ECA((C,\Fs,T), (C,\Fs^\prime,T^\prime)).
\end{align*}
Next, Equation~\ref{eq:mpa-sdp} can be derived as follows:
\begin{align}
&\MPA(\Fs^\prime) \nonumber \\
&= \sum_{\fs^\prime} \max \Bigg(\sum_{\rs} \left[C_T(\fs^\prime\rs)=c\right]\Pr(\rs\given\fs^\prime), \nonumber \\
&\phantom{= \sum_{\fs^\prime} \max \Bigg(} \sum_{\rs} \left[C_T(\fs^\prime\rs)=\overline{c}\right]\Pr(\rs\given\fs^\prime) \Bigg)\cdot\Pr(\fs^\prime) \label{eq:mpa-alt} \\
&= \sum_{\fs^\prime} \max \Big(\SDP_{C,T,0}(\Rs\given\fs^\prime), \nonumber \\
&\phantom{= \sum_{\fs^\prime} \max \Big(} 1-\SDP_{C,T,0}(\Rs\given\fs^\prime)\Big)\cdot\Pr(\fs^\prime). \label{eq:mpa-sdp2}
\end{align}
Equation~\ref{eq:mpa-alt} is obtained by simply factoring $\Pr(\fs^\prime\rs) = \Pr(\rs \given \fs^\prime)\cdot\Pr(\fs^\prime)$ from Definition~\ref{def:mpa}.
The first quantity in the maximizer term of Equation~\ref{eq:mpa-alt} is the expected probability of positive classification after observing features in $\Rs$ given that we already observed $\fs^\prime$.
This is equal to $\SDP_{C,T,0}(\Rs\given\fs^\prime)$, the same decision probability given that we always make a positive classification after $\fs^\prime$ (equivalent to using $T^\prime=0$). Thus, we get Equation~\ref{eq:mpa-sdp2} which defines the MPA using SDP.

Lastly, Equation~\ref{eq:eca-alt} follows from the connection of ECA to E-SDP in Proposition~\ref{prop:eca-esdp}, rewritten by partitioning the feature instances $\fs^\prime$ into those that lead to positive classification by $\beta$ and those that lead to negative classification.
Note that we can write $\SDP_{C,T,0}$ instead of $\SDP_{C,T,T^\prime}$ for the first sum because they equally measure the probability that additional observations will stick with positive classification.
For the second sum, we use the complement of $\SDP_{C,T,0}$ to express the probability that additional observations will lead to negative classification.

% }
\end{document}